\documentclass[conference]{IEEEtran}
\IEEEoverridecommandlockouts
\usepackage{cite}
\usepackage{amsmath,amssymb,amsfonts}
\usepackage{soul}
\usepackage{algorithmic}
\usepackage{graphicx}
\usepackage{textcomp}
\usepackage{xcolor}
\usepackage{amsfonts}
\usepackage{booktabs}
\usepackage{float}
\usepackage{graphicx}
\usepackage{grffile}
\usepackage{longtable}
\usepackage{wrapfig}
\usepackage{gensymb}
\usepackage{textcomp}
\usepackage{multicol}
\usepackage{amsmath}
\usepackage{algorithm}
\usepackage{verbatim}
\usepackage{graphicx}
\usepackage{caption}
\usepackage{listings}
\usepackage{array}
\usepackage{subcaption}
\usepackage{booktabs}
\usepackage{amsmath}
\usepackage{enumitem}
\usepackage{dblfloatfix}
\usepackage{balance}
\usepackage{lastpage}
\usepackage{lipsum}
\def\arraystretch{1.25}
\usepackage[utf8]{inputenc}

\title{Detecting Anomalous User Behavior in \\ Remote Patient Monitoring}

\author{\IEEEauthorblockN{Deepti Gupta\IEEEauthorrefmark{1}, Maanak Gupta\IEEEauthorrefmark{2}, Smriti Bhatt\IEEEauthorrefmark{3}, and Ali Saman Tosun\IEEEauthorrefmark{4} }
\IEEEauthorblockA{\IEEEauthorrefmark{1}\IEEEauthorrefmark{4}Dept. of Computer Science,
University of Texas at San Antonio,
San Antonio, Texas 78249, USA \\\IEEEauthorrefmark{2}{Dept. of Computer Science},
{Tennessee Technological University},
Cookeville, Tennessee 38505, USA \\\IEEEauthorrefmark{3}Dept. of Computing and Cyber Security, Texas A \& M University-San Antonio,
San Antonio, Texas 78224, USA\\}
\IEEEauthorrefmark{1}deepti.mrt@gmail.com, 
\IEEEauthorrefmark{2}mgupta@tntech.edu,
\IEEEauthorrefmark{3}sbhatt@tamusa.edu,  \IEEEauthorrefmark{4}ali.tosun@utsa.edu}

\begin{document}

\maketitle
\begin{abstract}
The growth in Remote Patient Monitoring (RPM) services using wearable and non-wearable Internet of Medical Things (IoMT) promises to improve the quality of diagnosis and facilitate timely treatment for a gamut of medical conditions. At the same time, the proliferation of IoMT devices increases the potential for malicious activities that can lead to catastrophic results including theft of personal information, data breach, and compromised medical devices, putting human lives at risk. IoMT devices generate tremendous amount of data that reflect user behavior patterns including both personal and day-to-day social activities along with daily routine health monitoring. In this context, there are possibilities of anomalies generated due to various reasons including unexpected user behavior, faulty sensor, or abnormal values from malicious/compromised devices. 
To address this problem, there is an imminent need to develop a framework for securing the smart health care infrastructure to identify and mitigate anomalies.
In this paper, we present an anomaly detection model for RPM utilizing IoMT and smart home devices. We propose Hidden Markov Model (HMM) based anomaly detection that analyzes normal user behavior in the context of RPM comprising both smart home and smart health devices, and identifies anomalous user behavior. We design a testbed with multiple IoMT devices and home sensors to collect data and use the HMM model to train using network and user behavioral data. Proposed HMM based anomaly detection model achieved over 98\% accuracy in identifying the anomalies in the context of RPM. 
 
\end{abstract}

\section{Introduction and Motivation}
\label{sec:introduction}
Internet of Medical Things (IoMT), also known as health care Internet of Things (IoT), is a critical data-driven application utilizing smart connected devices, relevant in context of COVID-19 like pandemic. This domain represents a connected infrastructure of medical devices, software applications, health care systems and services. There is an exponential growth in the number of IoMT devices and applications with utilization and demand across a diverse user population. According to the Grand View Research\footnote{https://www.healthcareitnews.com/news/asia-pacific/opportunities-pitfalls-healthcare-iot}, IoMT is predicted to reach \$534.3 Billion in market size by 2025. It is expected that the U.S. home-based healthcare market\footnote{https://store.businessinsider.com/products/the-us-home-healthcare-report} will rise by about 7\% annually from \$103 billion in 2018 to \$173 billion by 2026. 
With recent technological advancements, IoMT has the potential to revolutionize the future of health care industry. IoMT aims to escalate progress of the health care industry and enable people to receive timely care, enhance their treatment plans, manage medications, and lower health care costs. 

In the last few years, data-driven IoMT applications using Machine Learning (ML) and AI technologies are extensively used for critical functions, such as drug discovery, Remote Patient Monitoring (RPM), predictive analytic for hospital resource optimization, interactive medicine delivery, etc., by providing the right information at the right time in the health care system. Today, RPM has became a propitious and required need for users and health practitioners where patients can be monitored remotely, such as home isolation/quarantine. 
This RPM ecosystem consists of smart medical devices, such as thermometer, oximeter, and wearable devices.
These smart health devices collect a large amount of user health data. However, there are several issues associated with smart health devices and applications, such as security of devices and applications, privacy of user data which is shared between devices and applications among many. A recent report by the US National Institute of Standards and Technology (NIST)~\cite{cawthra1800securing} highlights the need on securing RPM ecosystem and discusses how health care delivery organizations can implement security and privacy controls to build a secure infrastructure for RPM.
 
In this paper, we focus on security of RPM including user safety and identify anomalous behavior in RPM applications based on the data collected from smart devices while remotely monitoring the patient at home. 
Research has developed various methods to secure RPM infrastructure from anomalies, focusing on wireless sensor networks~\cite{salem2013anomaly, vippalapalli2016internet,luo2012design}, patient's behavior monitoring~\cite{rachakonda2020ilog,bi2016anomaly,deep2019survey, swaroop2019health, gupta2020future}, signature and correlation analysis~\cite{jurdak2011wireless,fuhawatcher}. 
While there are several anomaly detection models developed for RPM, elderly care, and smart homes, our anomaly detection model is designed to identify anomalies in a unique RPM ecosystem which comprises the intersection of two IoT domains (smart health care and smart home). 
Our model aims to solve the problem of anomalous device data and user behavior which may occur due to malicious or faulty IoT devices or critical conditions of a user, for instance, when a user loses consciousness due to a heart attack and needs immediate medical attention. Proposed anomaly detection can ensure resilient RPM, and save a patient's life. 
 
A general RPM environment includes only medical devices, which makes it challenging to identify and differentiate  normal and abnormal data with user behavior. For instance, a smart oximeter reports the oxygen level of a diabetic patient as 70\%, which is very lower than normal value, thus, it will identify this as an anomaly. However, it could be a false positive due to faulty oximeter. In order to accurately identify anomalies in RPM, both data from medical devices and user activity behavior need to be monitored from home sensors in a smart home. In this work we design a novel RPM ecosystem with both medical and smart home devices that can monitor user health and daily activities within home, and detect anomalies accurately based on the correlation among user behavior and health readings. Extending the above scenario before labeling the oximeter reading as an anomaly, we need to consider these questions: a) Is the user normally moving at home? b) Is the user able to do normal tasks and respond to any alerts or suggestions on smart devices (e.g., smart watch)? and c) Is the user even at home when the oximeter reported 70\% oxygen level for the user? Smart home devices capture user's activities, for instance, after reported low oxygen level of the diabetic patient, if motion sensors at home detect no movement of the patient for certain time, then it will identify it accurately as an anomaly and will send an alert to health practitioner for remedial measures. 

In this paper, we develop an anomaly detection model using Hidden Markov Model (HMM) \cite{rabiner1989tutorial} approach to identify anomalies in the RPM environment that comprises of smart home and medical devices. We develop the RPM use case using our hybrid approach, and deploy it in a real-world scenario to train and test the model.  
Using our proposed model, health practitioners can identify anomalies in RPM and trigger appropriate actions as needed in a particular situation. Our model simulation results show that our proposed model detects anomalies in RPM with high accuracy over 98\% and generate low false positive rate.

The main contributions of this paper are as follows.
\begin{itemize}
\item We develop a novel use case of Remote Patient Monitoring (RPM) based on the intersection of two major IoT domains - smart health care and smart home. 
\item We design a testbed using sensors and smart health devices to implement the proposed RPM use case. 
\item We aggregate both network and user behavioral data collected from smart devices and convert it into temporal sequences for our anomaly detection model.  
\item We design an anomaly detection model using Hidden Markov Model (HMM) and test it based on aggregated training data that represents patient's normal behavior. 
\item We evaluate our proposed model on testing data along with sets of generated anomalies which achieved over 98\% accuracy to identify anomalies.
\end{itemize}
\begin{figure*}[t]
\centering
\includegraphics[width=1\textwidth, height = .27\textheight]{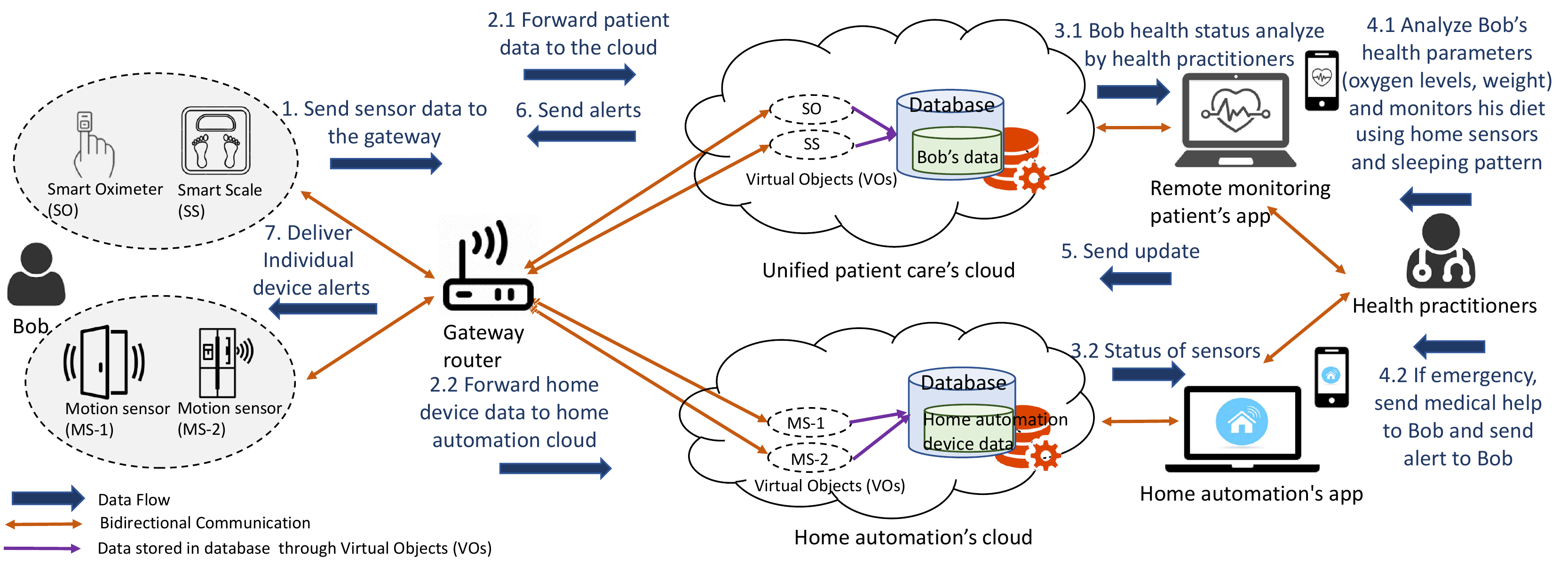}
\centering
\caption{A Use Case of Remote Patient Monitoring (RPM)}
\label{fig:RPM}
\end{figure*}

The remainder of this paper is organized as follows. Section~\ref{related} presents the literature review on anomaly detection models and machine learning approaches such as HMM. Section~\ref{sec:usecase} discusses the RPM use case and illustrate steps to develop the RPM model architecture. Section~\ref{sec:data} discusses network and behavioral data collection. Section~\ref{sec:model} discusses anomalous behavior detection model using HMM approach, and deployment for RPM use case. Section~\ref{sec:results} presents the implementation details and results, followed by conclusion in Section~\ref{sec:conclusion}.
\section{Related Work}
\label{related}
We present related work in anomaly detection models based on network and behavioral data.

\subsection{Anomaly Detection Models for RPM}
Extensive research has been done to keep RPM systems secure. Yamauchi et al.~\cite{yamauchi2020anomaly} introduced a method to detect attacks due to home appliances based on user behavior along with conditions at smart home and also presented a comparative study with HMM. Zhang et al.~\cite{zhang2013medmon} proposed a medical security monitor (MedMon) that snoops on all the radio-frequency wireless communications to/from medical implantable devices to identify anomalies. Tonchev et al.~\cite{tonchev2016non} presented a non-intrusive sleep analyzer for real time detection of sleep anomalies based on HMM and Viterbi algorithm. The authors used non-intrusive sensors such as bed pressure sensor, microphone, and accelerometer. Pham et al. \cite{pham2017wearable} used spectral coherence
analysis for accelerometer data, which is collected from wearable devices to develop an anomaly detection model. 
Deep et al.~\cite{deep2019survey} presented a survey on anomalous behavior detection for elderly case using dense-sensing networks and concluded that sensor fusion techniques could increase the efficiency of dense sensing network. Another study~\cite{parvin2018anomaly} showed that anomaly detection model for Ambient Assisted Living (AAL) focused on user's activities, such as sit down for dinner, open stoves etc., not related to user's body parameters. A study~\cite{ukil2016iot} presented an approach to reduce the false negative rate (FNR), i.e. disease should not get undetected. In order to protect the devices, security models are discussed in several research~\cite{gupta2020access, gupta2020secure, gupta2021towards, gupta2020learner, aslan2021intelligent}.

\subsection{Hidden Markov Models}
HMM models have been extensively used in various smart connected domains. HMM model-based behavior analysis system for assisted living is discussed in \cite{monekosso2010behavior}, where sensors are located in a smart home, each room with at least one motion detector, one temperature sensor, one light level detector and two lighting sensors status. 
In \cite{kang2010detecting}, a general framework is proposed for securing medical devices based on wireless channel monitoring and anomaly detection model using Hierarchical Hidden Markov Model (HHMM). Kotevska et al.~\cite{kotevska2019kensor} proposed coordinated intelligence approach to classify normal and abnormal behavior of resident using Markov chain at home where geographical co-located multiple sensors communicate with each other and take decisions based on the collective co-location in some capacity.
Li et al.~\cite{li2017multivariate} constructed HMM-based anomaly detectors and presented a comparative analysis based on several transformation methods. Research~\cite{ramapatruni2019anomaly} introduced an anomaly detection model based on user's behavior using HMM in a smart home environment. Narayanan et al.~\cite{narayanan2016obd_securealert} proposed OBD SecureAlert model for vehicles, which detects malicious behaviors and sends alerts while a vehicle is in unsafe state. Ren et al.~\cite{ren2017anomaly} introduced anomaly detection approach based on a dynamic Markov model where sequence data is managed by sliding window. Zhu~\cite{zhu2011automatic} used HMM approach for detecting anomalies in a person’s blood glucose levels using historical observations as a benchmark.


Extensive research has been done for anomaly detection in RPM and use of HMM in other smart domains for anomaly detection. However, as discussed earlier, an hybrid approach which uses both smart home and IoMT is still missing to detect anomalies more accurately (as suggested by our results). In addition, to our understanding and literature review HMM model has not been used in RPM. To bridge this gap, we believe our research offers a novel perspective to detect anomalies in RPM using HMM. 



\section{Remote Patient Monitoring (RPM)}
\label{sec:usecase}
In recent years, RPM has received significant attention due to its capability of providing health care services to patients while having the convenience of staying at home. Smart home-based health care and patient monitoring can enhance the quality of service and reduce the cost of health care.

\subsection{RPM Use case}

In this section, we present a RPM use case that is aligned with a real-world scenario. We consider a user \textit{Bob} who is 34 years old and lives alone at his home, and has been diagnosed with \textit{Obstructive Sleep Apnea (OSA)} disease. He is being monitored by his health practitioners continuously utilizing the RPM ecosystem enabled by smart devices deployed at his home. 
This RPM setup consists of set of devices including Ethernet tag manager, two wireless sensor tags, and various Bluetooth Low Energy (BLE) based smart health devices such as iHealth smart oximeter, iHealth smart scale. These specific smart health devices along with smart home devices are suggested by his health practitioners. Fridge door sensor and smart scale are used to monitor \textit{Bob}'s diet, eating behavior, and weight. Another motion sensor is attached on bedroom door to track \textit{Bob}'s activity, specially in night as \textit{OSA} patient has disturbed sleep and may wake up multiple times in night.
Smart oximeter keeps track of his oxygen level at regular intervals. In case \textit{Bob} is not able to sleep due to shortage of breathe, a notification is sent to the health practitioners based on data from bedroom door sensor and oxygen level through home automation cloud service and unified patient care cloud respectively. 
Lets us assume that the normal range of \textit{Bob}'s oxygen level is SpO\textsubscript{2} (88\% - 99\%). If \textit{Bob}'s oxygen level reaches less than 80\%, then the health practitioner is notified who can also send connected ambulance to \textit{Bob}'s home. 
Figure \ref{fig:RPM} shows a sequential view of the RPM use case scenario where \textit{Bob}'s health condition is continuously monitored by his health practitioners. \textit{Bob} checks his oxygen level and weight regularly in specific time intervals. These BLE based health devices communicate with a gateway device (\textit{Bob}'s smart phone) at object abstraction layer of Enhanced ACO architecture (EACO) \cite{bhatt2017access} to allow interaction with the cloud service and applications. 
The generated data from these devices is transmitted through the gateway to their corresponding manufacturer's cloud platforms (e.g., iHealth cloud and Wireless sensor cloud respectively) and is stored in  remote cloud. Users rely on the security mechanisms deployed by smart device manufacturers to ensure collected data privacy.

In general, anomalous behavior is defined as any violation or deviation from the normal behavior. We defined the anomalies that could occur in the RPM use case if a user deviates from his/her behavior or reflects any abnormal behavior. The following use case presents examples of abnormal behavior and possible threat scenarios.
\begin{itemize}
\item In RPM, the devices are used in a particular order based on user's daily routine. Any deviation could be an anomaly; for example, after checking oxygen level, a user opens the bedroom door within time window, if not that means user needs emergency assistance.
\item When user is not at home, but smart scale or oximeter are still sending data (vital readings). This could be due to a faulty smart device or possible intrusion at home.
\item An attacker can compromise an IoT device(s), for example, a user is measuring his oxygen level while other sensors like smart scale and fridge door are also activated at same time. 
\item Bedroom door sensor is activated multiple times in a short interval of time during night and the oxygen level of user is low,  i.e. user is not able to sleep during night. The condition can aggravate to a critical level and a health practitioner may need to intervene and check the patients health by sending messages to confirm his well-being. 
\end{itemize}



\subsection{System Architecture}

Our RPM model architecture is categorized into 3 phases:
\begin{itemize}
  
    \item \textbf{Data Collection Phase:} This is the first step in which the stream of data is collected from sensors and smart health devices to develop the anomaly detection model. 
   \item \textbf{Model Generation Phase}: We analyze the collected data and convert into sequences to feed the HMM model.
  \item \textbf{Anomaly Detection Phase:} In this final phase, anomalies are detected automatically using our proposed HMM-based anomaly detection model.

\end{itemize}

In these phases, we discuss and define normal behavior of a user and devices associated with that user as patterns in sequences of observations. For tracking the user behavior and training our anomaly detection model, \textit{opening the bedroom door, opening the fridge door, measuring weight and oxygen level} are the activities of our interest. Each activity and its relevant data is measured using corresponding smart devices in RPM ecosystem. These phases of RPM model architecture are described in the following sections.

\section{Data Collection}
\label{sec:data}
\begin{figure}[t]
\centering
\includegraphics[width=.5\textwidth, height =.3\textheight]{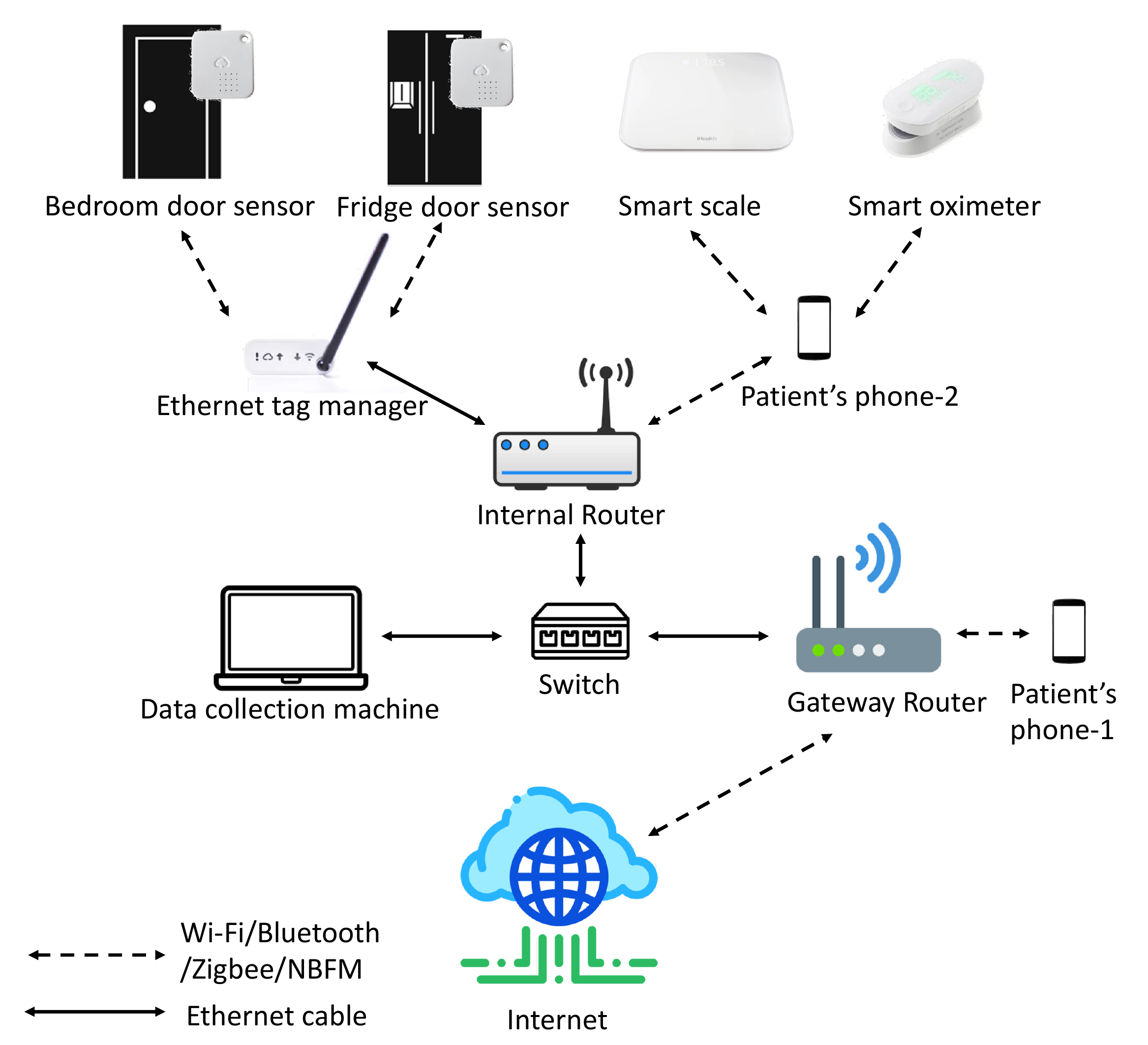}
\centering
\caption{Testbed Deployed for Remote Patient Monitoring}
\label{fig:setup}
\end{figure}

In RPM ecosystem, the smart connected devices generate tremendous amount of data while monitoring the patient at home. To collect this diverse data set, we set up the RPM testbed at a home by deploying IoT sensors and smart health devices (architecture shown in Figure \ref{fig:setup}). We consider that one patient lives in the smart home and performs his daily tasks, and the testbed captures data for three weeks. Data is collected from different sensors and smart health devices, which are used to build anomaly detection model.

We collect data in two different categories - (i) Network data, and (ii) Behavior data. We integrate both types of data to learn typical behaviors of a patient who uses smart medical devices and his activities are captured from home sensors. Network data refers to the TCP/UDP packets from IoT devices and smart phone. Behavioral data is collected based on the sensors' readings showing door status (open/closed) and health devices readings (e.g., 96\% SpO\textsubscript{2}). This behavioral data is fetched using API from different device manufacturer cloud data repositories (e.g., iHealth cloud, mytaglist cloud). 
\begin{figure*}[t]
\centering
\includegraphics[width=1\textwidth, height = 0.08\textheight]{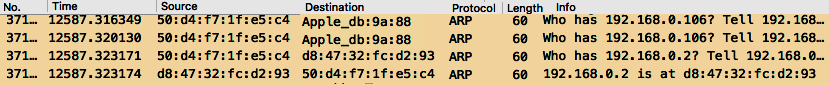}
\centering
\caption{Snapshot of Network Data Captured using Wireshark}
\label{fig:snap}
\end{figure*}

\subsection{Network Data Collection}
The varieties of data packet including audio and video flow through the gateway router of a smart home. However, we only require RPM related data from motion sensors and smart health devices. This required data cannot be separated from other non essential data inside the network, since the internal devices IPs are masked during Network Address Translation (NAT) in gateway router. Our setup consists of ARRIS SB6121 modem, two TP-Link AC1200 wireless router, NETGEAR ProSAFE Plus GS105Ev2 switch, and a data collection machine. The data collection machine is connected to the destination port of the switch and all data coming from the router will be available in it, as shown in Figure \ref{fig:setup}. The packets are collected at the switch using Wireshark\footnote{https://www.wireshark.org/} packet sniffer.  
We use the switch as a bridge between the gateway and internal router, and enable port mirroring in the switch. Port mirroring is a method of monitoring network traffic by setting up one or more source ports to send a copy of every packet received to a designated destination port. It is important to assign static IP to all devices in the testbed since IP address filtering is applicable if devices have a static IP. 

As shown in Figure \ref{fig:setup}, BLE based health devices are connected to patient's phone-2 to communicate its cloud service. To avoid other traffic, phone-2 is used only for data transfer of health devices and has set a static IP during three weeks of data collection. The motion sensors communicate to their cloud through Ethernet tag manager, and these wireless motion sensors use ``narrowband FM" technology\footnote{Long-range RF communication: Why narrowband is the de facto standard} to communicate to Ethernet tag manager. The static IP address of this tag manager is fixed via MAC address reservation.
\textit{Bob}'s phone-1 is used to keep track his presence at his home, where phone-1 connects to the gateway router. For instance, whenever user enters his home, phone-1 is connected Wi-Fi automatically. We filter the required network traffic of phone-1 to know \textit{Bob}'s presence at home, which is shown in Figure \ref{fig:snap}. This figure presents that phone-1 is connected to the gateway router, where mac address of gateway router is 50:d4:f7:1f:e5:c4 as source address and mac address of phone-1 is 20:ab:37:db:9a:88 as destination address. In this research, we define the status of phone-1 as `in' and `out' based on \textit{Bob}'s presence.



\begin{figure}[t]
\centering
\includegraphics[width=.41\textwidth, height = .14\textheight]{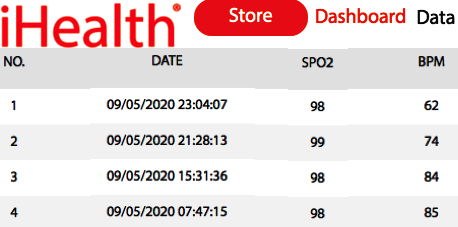}
\centering
\caption{Snapshot of iHealth Cloud Data}
\label{fig:sanp-2}
\end{figure} 


\subsection{Behavioral Data Collection}
The behavioral data represents the status of bedroom door, fridge door, and vital sign readings of smart health devices. In home, patient's movement is captured through sensors, for instance, if a motion sensor is attached to fridge door which identifies if a fridge door is open or close, i.e. patient is hungry or not. Another motion sensor is attached on bedroom door which identifies about sleeping time of patient. These motion sensors are attached on the doors with 30$^{\circ}$ angle to identify door is open or not. 
The home sensors are connected to Ethernet based smart tag manager, and this tag manager pushes the data into home automation's cloud. To differentiate between both sensors, the status of bedroom door are defined as `bd\_open' and `bd\_close', and the status of fridge door are defined as `fd\_open' and `fd\_close'. The status of the sensors and vital sign readings of health devices are retrieved from their corresponding clouds using API calls. Figure \ref{fig:sanp-2} shows body parameter readings for \textit{Bob} and from this data we derived the observation status for oximeter and smart scale. 
The smart health devices generate numeric reading, however, the devices’ numeric readings cannot be incorporated into logical calculations. We use normal distribution to discretize continuous data by using different thresholds, and calculate its mean $\mu$ and standard deviation $\sigma$ to define the range. For oximeter, three ranges (ox\textsubscript{1}, ox\textsubscript{2}, and ox\textsubscript{3}) are defined, where `ox\textsubscript{1}' comes under range of positive second deviation $\mu + 2\sigma$, which is 97\% to 99.3\%, and `ox\textsubscript{2}' comes under range of negative second deviation $\mu - 2\sigma$, which is 96.9\% to 94.7\%, if readings fall outside the range [$\mu - 2\sigma$, $\mu + 2\sigma$], is considered as `ox\textsubscript{3}'. The fourth status of oximeter is `ox\_off'.  We also define the range of smart scale in similar way. Patient's phone-2 has two status `on' and `off', and is connected to internal router and is only used to connect Bluetooth devices to send the data (vital readings) to iHealth cloud. Table \ref{tab:1} lists status of sensors and devices, which are considered as observations in HMM described in section \ref{sec:model}).

In our experiment, each sensor (door, fridge) sends its status every 30 seconds, while the status of IoMT devices (oximeter, scale) is measured by user at regular intervals. The user's presence is captured through his phone-1 through network traffic. To analyze, both network and behavioral data are integrated based on time-stamp, as shown in Table \ref{tab:2}. 
\begin{table}
\centering
\caption{List of Observations For Devices}
\label{tab:1}
\renewcommand{\arraystretch}{1.15}
\begin{tabular}{|c|c|} \hline
\textbf{Sensors/Devices}  & \textbf{Observations} \\ \hline \hline
Bedroom door & bd\_open, bd\_close\\ \hline
Fridge door & fd\_open, fd\_close \\ \hline
Weight scale & sc\_off, sc\textsubscript{1}, sc\textsubscript{2}, sc\textsubscript{3} \\ \hline
Oximeter &  ox\_off, ox\textsubscript{1}, ox\textsubscript{2}, ox\textsubscript{3} \\ \hline
Phone-2 & on, off \\ \hline
Phone-1 & in, out \\
\hline
\end{tabular}
\label{tab:Symbols}
\end{table}

\begin{table}[t!]
\centering
\caption{Sample log collected from the sensors in RPM}
\label{tab:2}
\renewcommand{\arraystretch}{1.15}
\begin{tabular}{|c|c|c|c|} \hline
\textbf{Time}  & \textbf{Sensors}  & \textbf{Status} & \textbf{Phone-2}\\ \hline \hline
8:02 & Phone-1 & in &  \\ \hline
8:04 & Oximeter & ox\textsubscript{1} & on \\ \hline
8:06 & Bedroom door & bd\_open & \\ \hline
8:07 & Bedroom door & bd\_close & \\ \hline
8:24 & Scale & sc\textsubscript{2} & on \\ \hline
8:32 & Fridge door & fd\_open & \\ \hline
8:33 & Fridge door & fd\_close & \\ \hline
\end{tabular}
\end{table}


\section{Anomalous Behavior Detection Model}
\label{sec:model}



In this section we propose an anomaly detection model for RPM ecosystem. We define the anomalies as abnormal activities performed by the user and abnormal values triggered from IoMT devices. This proposed model is based on HMM and considers the user behavior as specific sequences of data observations to feed the HMM model. 
We first present the HMM model, which learns the user's normal behavior and later define the proposed anomaly detection model for RPM use case scenario (discussed in Section \ref{sec:usecase}) .  

\subsection{HMM for Anomaly Detection in RPM}

We first collect diverse data from various IoMT devices and home sensors. Later, we analyze the collected data and convert into temporal sequences to feed HMM model. 
The HMM is an augmentation of Markov chain model, which is able to detect the sequential relations among hidden states. This is a probabilistic model where a sequence of observations is generated by visible observations of internal hidden states. These hidden states are not observed directly or are not visible. The transitions between hidden states are assumed to have the form of a (first-order) Markov chain. 

The specific parameters used to define the HMM are: $N$, $M$, $A$, $B$, and $\pi$, which are also mapped to the RPM use case. $N$ refers to the total number of all possible states in this model. At \textit{Bob}'s home, we consider his movement as a sequence of states, where sensors are activated at each state. Here, we denote an individual state from the set of states $S$ = ${S_1,S_2,S_3,....S_N}$, at time t as $q\textsubscript{t}$. 
$M$ is the number of unique visible observations for different sensors in a state, $V$ = ${v_1,v_2,v_3,....v_M}$ is the set of all possible observations. In our experiment, observations are simply status of different sensors, as shown in Table \ref{tab:1}. Parameter $A$ represents the state transition matrix. It gives the probability of being able to move from one state i to another state j, as described in equation 1. If a user does not move to any other state i.e. the transition probability would be zero. Parameter $B$ represents a probability distribution of seeing one of the observable status, given that user is in a particular state and is defined in equation 2. Finally, $\pi$ is the probability of starting at a particular state, which is randomly chosen. 

\noindent\begin{minipage}{.5\linewidth}
\begin{equation}
\scriptsize
\begin{split}
& A = [a\textsubscript{i,j}] \\
& a\textsubscript{i,j} = P(q\textsubscript{t+1} = S\textsubscript{j} | q\textsubscript{t} = S\textsubscript{i})\\
& where~q\textsubscript{t} \text{ is the state at time t,}\\
& S\textsubscript{i}, S\textsubscript{j} \in S(\text{set of all possible states})\\
& 1 \leq i \\
& j \leq N, t = 1,2,3,.......
\end{split}
\end{equation}
\end{minipage}%
\begin{minipage}{.5\linewidth}
\begin{equation}
\scriptsize
\begin{split}
& B = [b\textsubscript{i,j}] \\
& b\textsubscript{i,j} = P(v\textsubscript{k} \text{ at t} | q\textsubscript{t} = S\textsubscript{i})\\
& where~q\textsubscript{t} \text{ is the state at time t,}\\
& v\textsubscript{k} \in V (\text{set of all observations}) \\
& S\textsubscript{i}, S\textsubscript{j} \in S(\text{set of all possible states})\\
& 1 \leq i, j \leq N, t = 1,2,3,.......
\end{split}
\end{equation}
\end{minipage}

\subsection{Learning End User Behavior}
\begin{figure}[t]
\centering
\includegraphics[width=0.5\textwidth, height = .2\textheight]{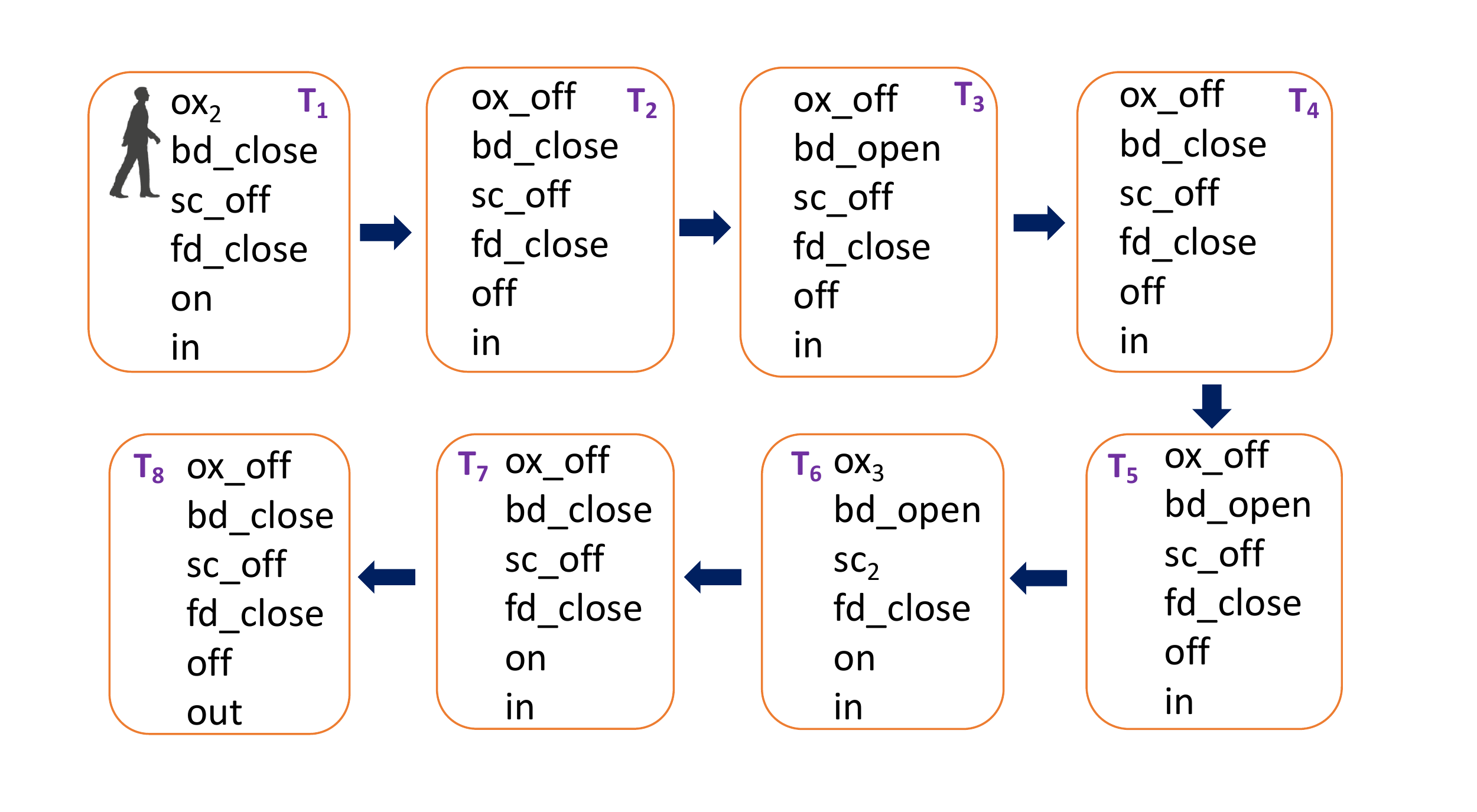}
\centering
\caption{Sample of Remote Patient Monitoring Events Timeline}
\label{fig:timeline}
\end{figure}

In RPM environment, \textit{Bob} uses medical IoT devices and his activities are captured through motion sensors. 
Figure \ref{fig:timeline} shows the timeline of \textit{Bob}'s movement at his home, as he moves from one state to another state, where multiple sensors and devices can be activated/deactivated at same time. 
At T\textsubscript{1}, oximeter is activated (observation `ox\textsubscript{2}') along with phone-2 (`on'), and at T\textsubscript{2}, the status of oximeter is changed from `ox\textsubscript{2}' to `ox\_off', and the status of phone-2 from `on' to `off'. Bedroom door is opened at T\textsubscript{3}, and close at T\textsubscript{4}. Then, bedroom door is open (bd\_open) again at T\textsubscript{5}, followed by multiple sensors (oximeter, bedroom door, scale, phone-2) activating at T\textsubscript{6}.
At T\textsubscript{6}, oxygen level is very low, bedroom door is also open, and smart scale is also activated. According to our model, the three possible sequences of observations will be generated from T\textsubscript{5} to  T\textsubscript{6}, e.g., [bd\_open, ox\textsubscript{3}], [bd\_open, sc\textsubscript{2}], and [bd\_open, off]. 
At T\textsubscript{7}, user is at home, phone-2 is still activated, while other sensors, devices are deactivated. At hidden state T\textsubscript{8}, status of phone is `off', user is out.


There are three issues in HMM which must be resolved using the following approaches. In one approach, given the sequence of observations O = O\textsubscript{1} O\textsubscript{2}...... O\textsubscript{T} and a model $\lambda$ = ($A$, $B$, $\pi$) is used to evaluate \begin{math}P(O\vert\lambda)\end{math} probability of any sequence of observations using Forward-Backward algorithm \cite{rabiner1989tutorial}. Next approach is used to adjust the model parameters $\lambda$ = ($A$, $B$, $\pi$) to maximize \begin{math}P(O\vert\lambda)\end{math} using Baum-Welch algorithm \cite{rabiner1989tutorial}. In our experiment, we collect time-series data from all the devices and sensors, and convert them into temporal sequences, which will be the training sequences of the model. 
The training sequences O = O\textsubscript{1}, O\textsubscript{2},... ,O\textsubscript{K} are extracted from this time-series data, where O\textsubscript{K} is an observation at each state. 
In our experiment, we assume that  O\textsubscript{K} is vector as multiple sensors are activated at same time (shown in Figure \ref{fig:timeline}), where O\textsubscript{K} = $v_{1}^{(k)}$ $v_{2}^{(k)}$..........$v_{t}^{(k)}$, 1$\leq$ k $\leq$ K. Observation vector (O\textsubscript{K}) implies to generate different combinations of sequences.
For example, sequence of observations extracted during state transitions from time T\textsubscript{1} to T\textsubscript{5} using Figure \ref{fig:timeline} is [\{ox\textsubscript{2},on\},\{ox\_off,off\}, bd\_open, bd\_close, bd\_open].
In our experiment, we extracted 420 sequences of observations from temporal data, and trained the model with these sequences using Baum-Welch algorithm to determine the HMM model parameters $\lambda$ = ($A$, $B$, $\pi$). The training process repeats until the resulting probabilities converge satisfactorily.
For testing, Forward-Backward algorithm is used to evaluate the new sequence of observations by determining the likelihood P(O$\vert$$\lambda$), which is the log probability of new sequence of observations.
The high value of log probabilities indicates that new sequence of observations fit for the model, and low value of log probabilities implies deviation from normal behavior, and identify as anomaly. 




\subsection{Anomaly Detection Model}
\begin{figure}[t]
\centering
\includegraphics[width=0.33\textwidth, height = .14\textheight]{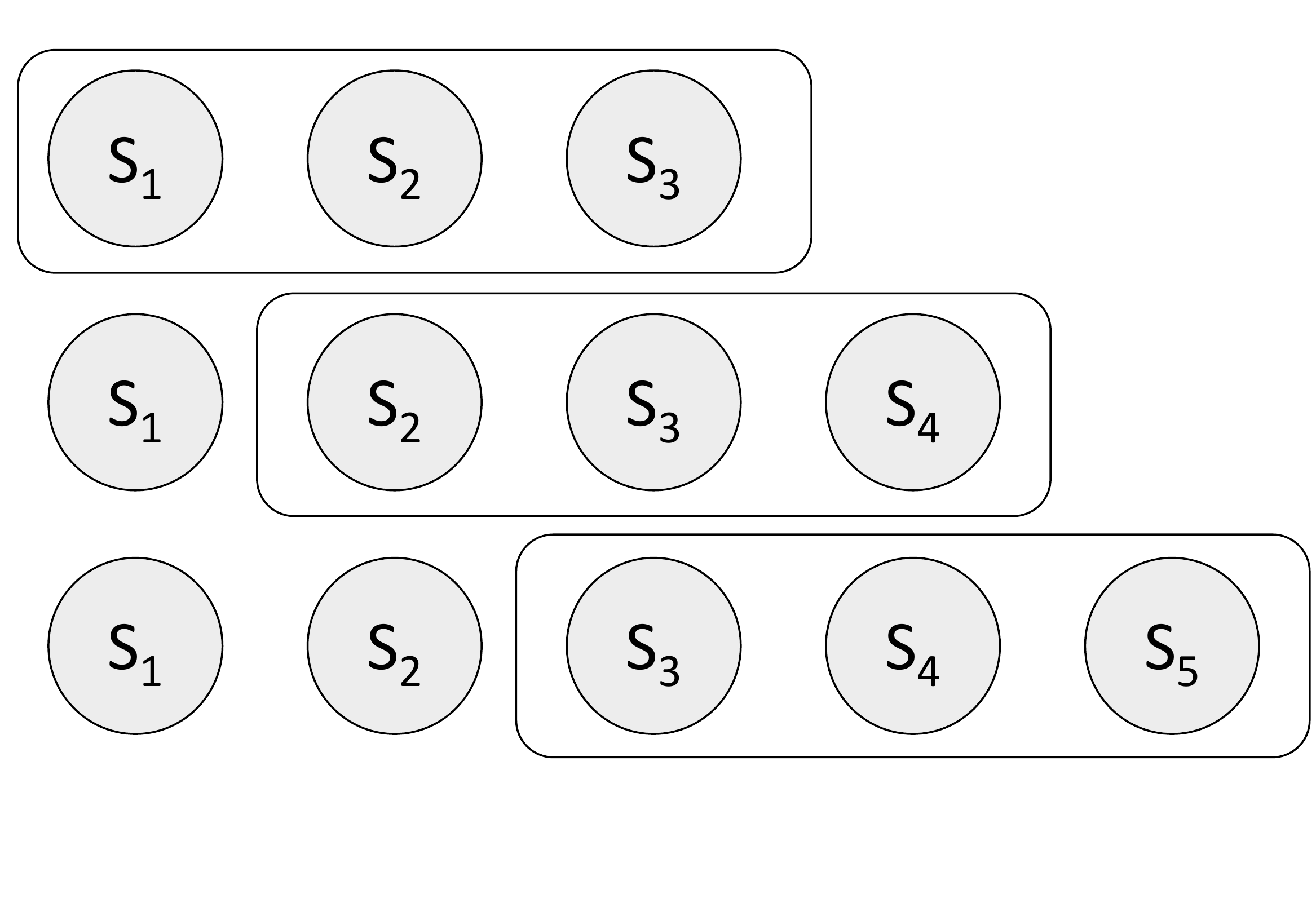}
\centering
\caption{Sliding window to detect anomalies using Forward-Backward Algorithm}
\label{fig:window}
\end{figure}

After generating transition, emission and initial probabilities matrices using Baum-Welch algorithm, we use this generated HMM model to detect anomalies in RPM environment. 
In this research, we are detecting both attack states and unsafe/anomalous states. Here is example of faulty sensors, when user is away from his home, and smart scale is still sending readings to patient's care cloud. To detect the anomalies, we generated a sliding window of hidden states as shown in Figure \ref{fig:window}, where observation vector O\textsubscript{K} is visible at each state. To detect anomalies in real-time, a sliding window of “m” observations, O\textsubscript{window} = {O\textsubscript{1},O\textsubscript{2},....O\textsubscript{m}} is introduced. The sliding window moves every time a new observation vector is available at a new state and calculate the log probability of new generated sequences of observations using Forward-Backward algorithm. As we define that each observation would be a vector of different sensor values, thus, it will generate a set of log probabilities corresponding to each sequence of observations. If any log probability in this set is below a threshold (discussed in Section \ref{sec:results}), it identifies it as an anomalous state.
Any sequence of observations is considered anomalous, if it is not accepted by the proposed model with low probability.



\section{Evaluation and Results}
\label{sec:results}

In this section, we present the evaluation of our proposed model and also discuss the results. We generated 420 sequences of observations from collected data during the period of three weeks. This data represents \textit{Bob}'s normal behavior in RPM environment. We divide collected data into training and testing datasets, and calculate the threshold value based on the lowest value of log probability of training sequences in various settings. The threshold value is used to identify the anomalies in RPM ecosystem. Further, we generate sets of anomalous scenarios to validate the proposed model. The details of the experiments are presented in the following subsections.

\subsection{Analysis of Normal Dataset}

\begin{figure}[t]
\centering
\includegraphics[width=0.5\textwidth, height = .13\textheight]{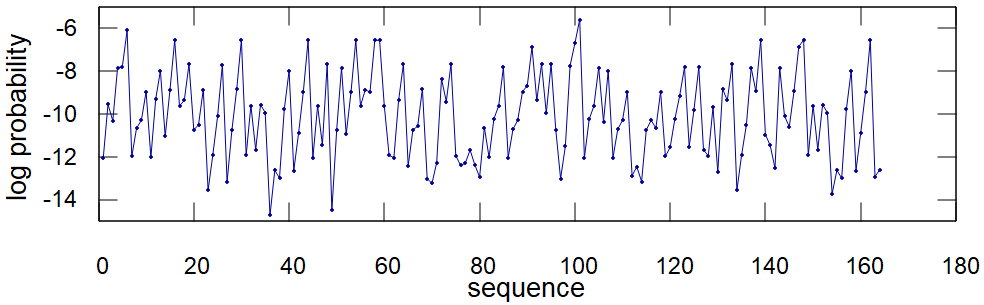}
\centering
\caption{Test data for Remote Patient Monitoring}
\label{fig:R-1}
\end{figure}

For training, we use Baum-Welch algorithm and use Forward-Backward algorithm for testing. It is required to generate a threshold value to test any sequence using Forward-Backward algorithm. To validate the threshold value, we divide the experimental data into three different settings and perform these experiments. For our first experiment, 70\% of data is used for training and 30\% is used for testing.
We train the model using 70\% sequences of observations and calculate the log probabilities of training sequences, which are in the range of -5 to -14. For second experiment, we train the HMM model with the first 60\% of data and observed that the log probability of training sequences ranges from -6 to -14. In third experiment, we divide 420 sequences into two parts, each set has 210 sequences, where 70\% sequences are used for both training data and provide log probability ranges (-5 to -14, -6 to -13). Hence, we set the threshold value to be -14, i.e. the log probability value of any sequence lower than -14 is anomaly. The testing sequences of above defined three experiments are tested on our proposed model using their corresponding training datasets, and this model provide different sets of log probability ranges (-5 to -14.9, -7 to -14, -6.5 to -13.4, -6 to -14).
The range of log probabilities of testing data of first experiment is shown in Figure~\ref{fig:R-1}, which also shows two false positives. 


\begin{figure}[t]
\centering
\includegraphics[width=0.5\textwidth, height = .18\textheight]{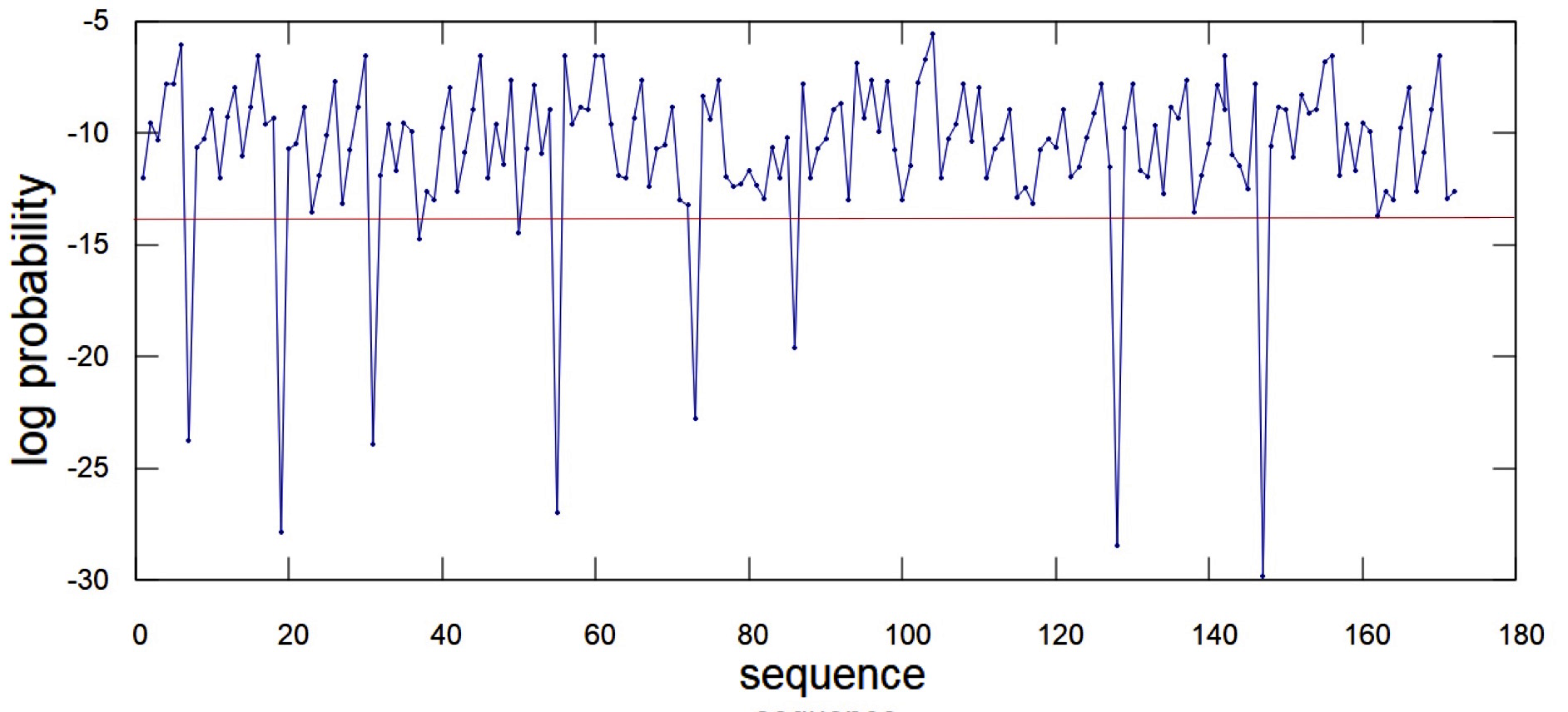}
\centering
\caption{Generated anomalies injected into test normal data}
\label{fig:R-2}
\end{figure}
\subsection{Abnormal Scenario Detection}
To evaluate the model's performance, we manually generate a set of eight anomalous scenarios which are representative of real-world scenarios. We consider abnormal behaviors including unusual device reading, missing dependency order between devices, and compromised sensors/devices to select eight anomalous scenarios. 
First anomalous scenario is \textit{User is out from his home, smart scale is still sending reading to the unified care's cloud}. It may happen due to faulty/compromised device, which can effect  \textit{Bob}'s diet. To simulate this scenario, we take a normal sequence of observation, activate the smart scale and modify the status of phone-1 from `in' to `out'. Another crafted scenario is \textit{SpO\textsubscript{2} is very low (i.e. ox\textsubscript{3}) and bedroom door is not activated.}, i.e. oxygen level of user is low, and he is not able to get up to open the bedroom door within a time frame. Third anomalous scenario is \textit{when user is out, scale, oximeter and fridge door sensor activate simultaneously}, that indicates the presence of another user at home or any malicious activity can be possible. 
Fourth and fifth anomalous scenarios are \textit{when user is at home, fridge door, oximeter and scale activate simultaneously}, and \textit{oximeter reading is reported as ox\textsubscript{2}, but ox\_off is not followed by ox\textsubscript{2} within a time frame} respectively.  Sixth anomalous scenario is \textit{when user is at home, sc\textsubscript{3} is followed by ox\textsubscript{3}}, i.e. only medical devices are activated with unusual vital readings and seventh scenario is \textit{when user is out of his home, bd\_open and fd\_open are activated}. The last anomalous scenario is \textit{bd\_open followed by bd\_open along with ox\textsubscript{3}}, i.e. due to low level of oxygen, user is not able to sleep.  
These eight generated anomalous scenarios are injected into normal testing data of first experiment and tested on proposed anomaly detection model. The threshold values of log probabilities of these eight anomalous scenarios are lower than -14, which is shown in Figure \ref{fig:R-2} along with two false positives. Hence, the proposed model detects all generated anomalies in RPM environment. 



In the next experiment, we generate another set of anomalous scenarios by fixing or randomly choosing status of specific sensors/devices. For example, we change the `ox\textsubscript{2}' to `ox\textsubscript{3}', `on' to `off', and `in' to `out'. In total, we generated 38 anomalous scenarios randomly, and added eight generated anomalous scenarios to create a dataset of 46 anomalous scenarios. This dataset is tested on our trained model, which detects 45 anomalies out of 46. Moreover, this model can work with any size of dataset. The confusion matrix of the first experiment (30\% of 420) along with the 46 anomalous scenario dataset is given in Table~\ref{tab:4}, which shows that our proposed HMM based approach provides 98\% accuracy.

\begin{table}
\centering
\caption{Confusion Matrix}
\label{tab:4}
\begin{tabular}{l|l|c|c|c}
\multicolumn{2}{c}{}&\multicolumn{2}{c}{}&\\
\cline{3-4}
\multicolumn{2}{c|}{N=172}&Actual:Yes&Actual:Yes&\multicolumn{1}{c}{Total}\\
\cline{2-4}
{ }& Predicted:Yes & $TP=45$ & $FP=2$ & $47$\\
\cline{2-4}
& Predicted:No & $FN=1$ & $TN=124$ & $125$\\
\cline{2-4}
\multicolumn{1}{c}{} & \multicolumn{1}{c}{Total} & \multicolumn{1}{c}{$46$} & \multicolumn{1}{c}{$126$} & \multicolumn{1}{c}{}\\
\end{tabular}
\end{table}
\section{Conclusion}
\label{sec:conclusion}

In this paper, we propose an anomaly detection model for the RPM ecosystem consisting of smart home and medical IoT devices. We design a testbed of RPM, collect time-series data, and convert data into sequences of observations which are used to train the HMM model using Baum-Welch algorithm. 
This proposed model detects anomalies based on a single device, and also identifies anomalies based on the combination of data collected from both IoMT and home-based devices. 
Our model evaluation shows that the proposed HMM based approach can detect anomalies in RPM real-world events/scenarios and randomly generated anomalies with 98\% accuracy. This model can be adopted for a diverse range of anomalous use cases. For future, we plan to develop a robust anomaly detection model for multi-users behavior with geographically co-located sensors in home-based health care applications with more anomalous events. 


\section*{Acknowledgement}
This research is partially supported by the NSF Grant 2025682 at Tennessee Technological University, and the Chancellor Research Initiative (CRI) grant at Texas A\&M University-San Antonio. The authors would like to thank Dr. Sudip Mittal for his suggestions. 

\balance
{
\bibliographystyle{IEEEtran}
\bibliography{References}
}
\end{document}